\documentclass[letterpaper, 10 pt, conference]{ieeeconf}  

\IEEEoverridecommandlockouts                              

\UseRawInputEncoding

\usepackage{graphics} 
\usepackage{epsfig} 
\usepackage{xcolor}
\usepackage{multirow}
\usepackage{longtable}
\usepackage{amsmath}
\usepackage{float}
\usepackage{url}
\usepackage{subcaption}
\usepackage{mathtools}
\usepackage{amsmath}
\usepackage{graphicx}
\usepackage{hyperref}
\usepackage{amssymb}
\usepackage{lipsum,lineno}
\usepackage[symbol]{footmisc}

\usepackage{booktabs}
\usepackage{multirow}
\usepackage{bbm} 
\let\labelindent\relax
\usepackage{enumitem}

\DeclareMathOperator*{\argmax}{argmax}

\newcommand{\sobj}[1]{s_{\text{o}_{#1}}}

\newcommand{\srobot}[0]{s_{r}}
\newcommand{\StateS}[0]{\mathcal{S}}

\newcommand{\abase}[1]{a_{\text{bp}}^{#1}}
\newcommand{\aobj}[1]{a_{\text{o}_{#1}}}
\newcommand{\aseq}[0]{\mathcal{A}_{\text{seq}}}

\newcommand{\ActionS}[0]{\mathcal{A}}

\newcommand{\pbase}[0]{\pi_{\text{bp}}}
\newcommand{\pseq}[0]{\pi_{\text{seq}}}

\newcommand{\rbase}[0]{R_{\text{bp}}}
\newcommand{\rseq}[0]{R_{\text{seq}}}
\newcommand{\rbaseline}[0]{R_{\text{baseline}}}

\newcommand{\tnav}[0]{t_{\text{nav}}}
\newcommand{\tgrasp}[0]{t_{\text{grasp}}}

\newcommand{\obj}[1]{o_{#1}}
\newcommand{\robot}[0]{r}

\newcommand{\world}[0]{\mathbf{W}}

\newcommand{\objset}[0]{\mathcal{O}=\{\obj{n}\}_n}

\newcommand{\transform}[2]{\mathbf{T}_{#2}^{#1}}

\newcommand{\name}[0]{\textsc{BaSeNet}}

\title{\LARGE \bf \name: A Learning-based Mobile Manipulator Base Pose Sequence Planning for Pickup Tasks}

\author{Lakshadeep Naik$^{1}$, Sinan Kalkan$^{2}$, Sune L. S\o rensen$^{3}$, Mikkel B. Kj\ae rgaard$^{3}$, and Norbert Kr\"uger$^{1,4}$
\thanks{$^{1}$SDU Robotics, M\ae rsk Mc-Kinney M\o ller Institute (MMMI),  Faculty of Engineering, University of Southern Denmark, Odense M, Denmark {\tt\small \{lana,norbert\}@mmmi.sdu.dk}}%
\thanks{$^{2}$Department of Computer Engineering, Middle East Technical University, Ankara, Turkey {\tt\small skalkan@metu.edu.tr}}%
\thanks{$^{3}$SDU Software Engineering, M\ae rsk Mc-Kinney M\o ller Institute (MMMI),  Faculty of Engineering, University of Southern Denmark, Odense M, Denmark {\tt\small \{slso,mbkj\}@mmmi.sdu.dk}}%
\thanks{$^{4}$Danish Institute for Advanced Studies (DIAS), Odense M, Denmark}%
}
\begin{document}

\maketitle
\thispagestyle{empty}
\pagestyle{empty}


\begin{abstract}
In many applications, a mobile manipulator robot is required to grasp a set of objects distributed in space. This may not be feasible from a single base pose and the robot must plan the sequence of base poses for grasping all objects, minimizing the total navigation and grasping time. This is a Combinatorial Optimization problem that can be solved using exact methods, which provide optimal solutions but are computationally expensive, or approximate methods, which offer computationally efficient but sub-optimal solutions. Recent studies have shown that learning-based methods can solve Combinatorial Optimization problems, providing near-optimal and computationally efficient solutions. 

In this work, we present \name~- a learning-based approach to plan the sequence of base poses for the robot to grasp all the objects in the scene. We propose a Reinforcement Learning based solution that learns the base poses for grasping individual objects and the sequence in which the objects should be grasped to minimize the total navigation and grasping costs using Layered Learning. As the problem has a varying number of states and actions, we represent states and actions as a graph and use Graph Neural Networks for learning. We show that the proposed method can produce comparable solutions to exact and approximate methods with significantly less computation time. The code, Reinforcement Learning environments, and pre-trained models will be made available on the project webpage\footnote[1]{\url{https://lakshadeep.github.io/basenet/}}.
\end{abstract}

\section{INTRODUCTION}
\label{sec:intro}
Mobile Manipulators (MMs) are widely used for pick-up
tasks across different domains including logistics, manufacturing, service, home automation, elderly care, and hospitality applications \cite{roa2021mobile, asgharian2022review}. Pickup tasks involve determining suitable robot base poses for object pick-up, followed by navigating to the base pose and picking up the objects using the manipulator \cite{sandakalum2022motion, reister2022combining}.

\begin{figure}[hbt!]
    \centering
    \includegraphics[width=1.00\columnwidth]{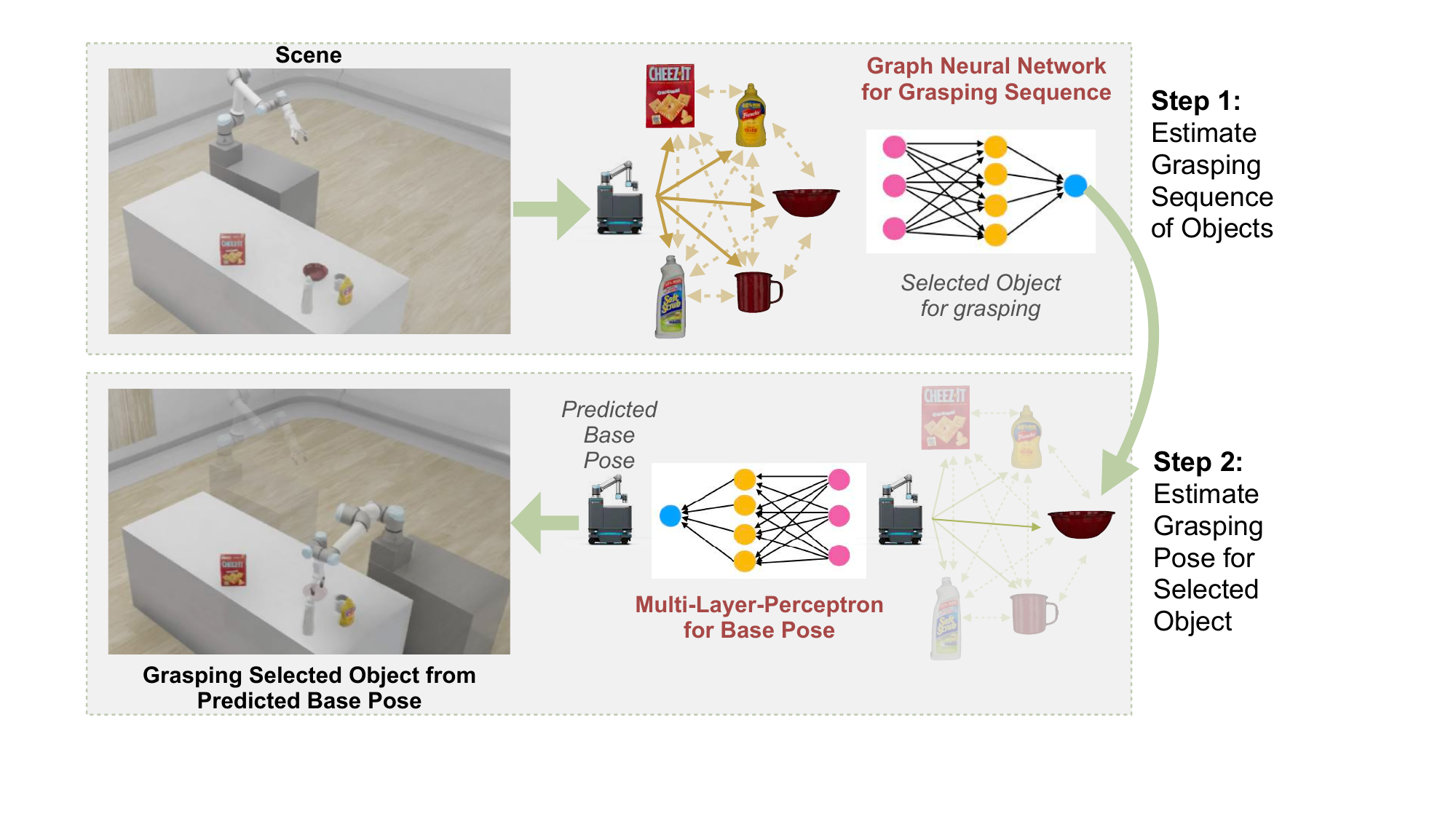}
    \caption{\name~ \textbf{Top row:} The robot represents the scene as a graph and determines the next object to grasp using the \textit{grasp sequence} policy.
    \textbf{Bottom row:} The robot predicts the base pose for grasping the selected object using the \textit{base pose} policy and performs the action.}
    \label{fig:idea}
\end{figure}

In challenging environments, multiple base poses are generally required for grasping all the objects (see Fig.~\ref{fig:idea}). In such situations, the robot must plan the optimal sequence of base poses for picking up the objects such that the total navigation and grasping time is minimized. This is a Combinatorial Optimization (CO) problem \cite{papadimitriou1998combinatorial}, which can be addressed using exact methods such as dynamic programming, which ensure optimal solutions \cite{held1962dynamic, sorensen2024planning}. However, these methods are computationally expensive \cite{han2018complexity, wang2021optimal}, prohibiting their use in practice on robots as they often require re-planning due to changes in the object configuration, such as those caused by human interference or collisions with other objects during the picking process. Consequently, various approximate solutions \cite{wang2021optimal, diankov2010automated, du2012optimal, harada2015base, vafadar2018optimal, xu2020planning, zhang2023base} have been proposed, balancing optimality and computational efficiency \cite{kool2018attention}.

Several recent works have explored using learning-based methods for solving CO problems, such as routing problems \cite{mazyavkina2021reinforcement, cappart2023combinatorial}. These methods offer near-optimal and computationally efficient solutions. Drawing inspiration from these works, we propose a learning-based approach for determining the optimal sequence of base poses for grasping all the objects in the scene. 

However, a significant distinction exists between the routing problem and determining the sequence of base poses for grasping. In the routing problem, costs depend on the node features (for example, city coordinates in the Travelling Salesman Problem). In the base pose sequence planning problem, node features consist of object poses and the cost depends on the base pose selected for grasping the object. Furthermore, each object can be grasped from several different base poses. Thus, in addition to learning the optimal sequence in which the objects should be grasped, the base pose for grasping each object also needs to be learned. This makes it more challenging to learn compared to routing problems. Furthermore, due to the limited view of the robot's onboard camera and uncertainty in the robot's self-localization, often only uncertain object poses are available for base pose sequence planning.

Existing works that learn base poses for grasping have focused on grasping a single object \cite{jauhri_robot_2022, naik2024}.
For planning the sequence of base poses for grasping multiple objects, the current robot pose as well as the poses of all the objects in the scene also must be considered. This poses two main challenges:
\begin{enumerate}
\item \textit{Varying number of states and actions.} As the number of objects in the scene can vary, the state and actions cannot be represented using a fixed-dimensional vector.
\item \textit{Sample inefficiency.} Learning in such a high-dimensional state and action space requires a large amount of training data.
\end{enumerate}
We address the first challenge by representing states and actions as a graph and using Graph Neural Network (GNN) to encode a state into a fixed dimensional vector. To address sample inefficiency, we use Layered Learning (LL) \cite{stone2000layered} in combination with Reinforcement Learning (RL) similar to our previous work \cite{naik2024}. In Layer 1, we learn the \textit{grasp sequence} policy which selects the next object to grasp among the remaining objects (see Fig.~\ref{fig:idea} first row). In Layer 2, we learn the \textit{base pose} policy which predicts the base pose for grasping the object selected by \textit{grasp sequence} policy (see Fig.~\ref{fig:idea} second row). We choose to ignore object pose uncertainties to simplify learning. Moreover, by augmenting robot onboard camera views with external cameras in the environment and temporal fusion, accurate pose estimates can be obtained for pre-grasp planning \cite{naik2022multi} such as the base pose sequence planning.

To summarize, we make the following contributions:
\begin{enumerate}
\item We formulate the problem of base pose sequence planning to optimize total navigation and grasping costs as an RL problem.
\item We sequentially learn the base poses for grasping individual objects and the object grasp sequence using LL.
\item We address the variable state and action space challenge in grasp sequence planning by formulating the problem as a graph node regression problem.
\item Through experimental evaluation, we show that \name~ can reduce the total planning and execution time by more than 50\% compared to the best-performing baselines with almost the same success rate.
\end{enumerate}


\section{RELATED WORK}
\label{sec:related_work}

\subsection{Explicit base pose planning} 
\label{ss:rw:bpp}

The selection of a base pose for grasping an object relies on the availability of valid Inverse Kinematics (IK) solutions to achieve the desired grasp pose. Searching for base poses with valid IK solutions in SE(2) can be computationally intensive. Therefore, existing works have suggested the utilization of Inverse Reachability Maps (IRM) \cite{makhal2018reuleaux,vahrenkamp2013robot}. IRM discretizes the base pose space using a grid-based approximation and stores the base poses from which IK solutions are available for the selected grasp pose in the offline phase. During online execution, a specific heuristic is employed to select a particular base pose. The availability of an IK solution does not guarantee that a valid trajectory can be planned to the desired end-effector pose, as trajectory planning depends on several factors such as self-collision, collision with other objects in the scene, joint limits, manipulability ellipsoid of the manipulator, etc. As a result, additional online validations are required to ensure that the trajectory can be planned from the selected base pose.

Recent works have also proposed learning-based methods to predict the base pose for grasping single objects \cite{jauhri_robot_2022,kim2021learning}. These methods have shown to be much more computationally efficient and do not suffer from grid-based approximation like IRM.

\noindent{\textbf{Difference.}} In this work, we learn to plan the optimal base pose for grasping an object while also minimizing the combined cost of navigating to the selected base pose from the robot's current base pose and grasping the object.

\subsection{Grasp sequence planning}
\label{ss:rw:gsp}
Being a Combinatorial Optimization (CO) problem, grasp sequence planning can be solved using exact methods that provide optimal solutions at high computational cost \cite{sorensen2024planning}, or evolutionary \cite{berenson2008optimization} or heuristic methods \cite{wang2021optimal,diankov2010automated, du2012optimal, harada2015base, vafadar2018optimal, xu2020planning} that offer sub-optimal solutions at low computational cost.

S{\o}rensen et al. \cite{sorensen2024planning} have employed dynamic programming with memoization to find optimal base pose sequences; however, the quality of obtained solutions directly depends on the action space resolution used for computing the costs. High action space resolution for cost computation produces better solutions but at a high computational cost. Most works that find sub-optimal but quick solutions using non-exact methods utilize IRM and make certain assumptions, such as all objects can be grasped from a single base pose \cite{berenson2008optimization, du2012optimal}, or base pose orientation is fixed \cite{harada2015base, xu2020planning}, or the order in which the objects should be grasped is already known \cite{reister2022combining}, to simplify the complexity of the problem.

\noindent{\textbf{Difference.}} In this work, instead of making any such assumptions, we let the robot itself explore the base pose space for grasping objects in the scene and learn the optimal base pose sequence.

\subsection{Combinatorial optimization and learning}
\label{ss:rw:col}
Exact methods, such as dynamic programming, can be applied to any generic CO problems to obtain optimal solutions, albeit at a very high computational cost. Conversely, approximate methods provide quick but sub-optimal solutions by making certain assumptions designed by domain experts to simplify the problem. Moreover, for similar problem instances, such as base pose sequence planning for the same workspace, the optimal solutions would be similar. Hence, learning techniques such as RL can be employed to search for heuristics using data instead of hand-crafted heuristics \cite{mazyavkina2021reinforcement}.

Initial works with learning-based solutions for CO problems, such as Pointer networks \cite{vinyals2015pointer}, used supervised data to find the solutions. Later works, such as \cite{kool2018attention,bello2016neural}, trained policies in an unsupervised manner using RL, attention mechanisms \cite{vaswani2017attention}, etc. In recent years, Graph Neural Networks (GNN) \cite{wu2020comprehensive} have emerged as efficient state representations for CO problems. GNNs can learn the vector representation that encodes crucial graph structures required to solve CO problems efficiently \cite{cappart2023combinatorial}.

\noindent{\textbf{Difference.}} In this work, we employ the Graph Attention Layers \cite{velivckovic2018graph} to learn a vector representation that encodes relevant grasp scene information for learning the grasp sequence in an unsupervised manner using REINFORCE with greedy rollout baseline, similar to \cite{kool2018attention}.


\section{PROBLEM FORMULATION}
\label{sec:pf}

We address the problem of picking up a set of $N$ rigid objects $\objset$ from a table using a mobile manipulator robot. We assume that the objects can be grasped using an overhead (top-down) grasp and that the robot has a navigation stack \cite{guimaraes2016ros} to navigate to the planned base pose and a manipulation stack \cite{chitta2012moveit} for grasping.

It may not be possible for the robot to pick up all objects from one base pose $\abase{n}$ ($\in \mathrm{SE(2)}$) and hence may have to move to a sequence of base poses 
\begin{equation}
    \aseq = \{ \abase{1}, \abase{2}, ...\abase{N} \},
\end{equation}
to pick up $N$ different objects. Our objective is to determine the optimal sequence for grasping objects and the corresponding base pose for each object, ensuring time-efficient completion of the pickup task.

We formulate this as an RL task. During the training stage, $N$ objects $\objset$ are randomly placed on the table. Each RL episode consists of a maximum of $N$ steps. At each step, the robot predicts the next base pose $\abase{n}$ ($n\in [1, N]$) and the object to grasp $\obj{m}$. The objective of the RL agent is to complete the task efficiently, minimizing the total execution time for navigation $\tnav$ and grasping $\tgrasp$.

Learning such a policy requires information about the current robot base pose and the object poses. Thus, the state space consists of: 
\begin{equation}
    \StateS = \{ \srobot, \{ \sobj{1}, \sobj{2}, ...\sobj{M} \}\},
\end{equation}
where $\srobot \in \mathrm{SE(2)}$ represents the robot base pose in the table frame $\world$, $\sobj{m} \in \mathrm{SE(2)}$ denotes the $m$-th object pose in the table frame $\world$, and $M \leq N$ is the number of objects yet to be grasped. The action space consists of actions:
\begin{equation}
    \ActionS = \{ \abase{n}, \{ \aobj{1}, \aobj{2}, ... \aobj{M}\} \},
\end{equation}
where $\abase{n} \in \mathrm{SE(2)}$ represents the predicted next robot base pose in the frame of the object selected for grasping, and $\aobj{m}$ signifies the probability of grasping object $\obj{m}$.

An episode ends when the agent exceeds $N$ steps, collides with the table, or when all the objects are grasped. Further, all states and actions are internally represented with a time variable $t$, which, however, is omitted in our notations for convenience.

\begin{figure*}[h]
    \centering
    \includegraphics[width=17.5cm]{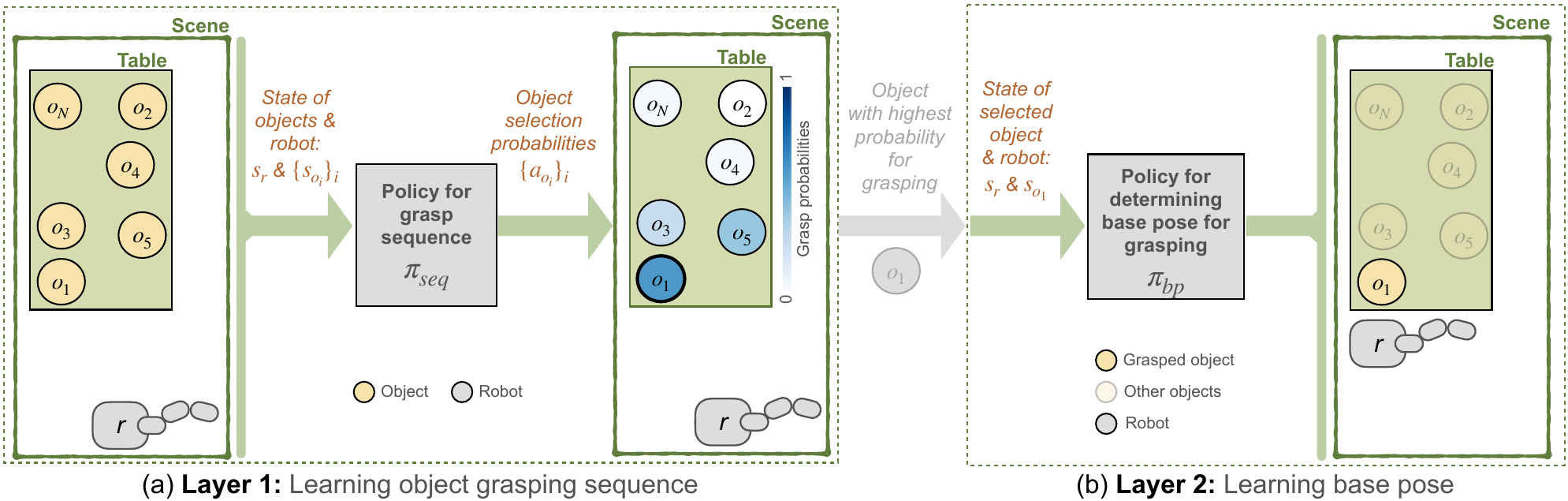}
    \caption{Proposed Layered Approach - \name: \textbf{(b) Layer 2:} Learning base poses for grasping individual objects (\textit{base pose} policy $\pbase$). \textbf{(a) Layer 1:} Learning object grasping sequences (\textit{grasp sequence} policy $\pseq$) using already learned $\pbase$ for determining the base pose for grasping for the selected object. The example demonstrates that in Layer 1, the object $o_1$ is selected for grasping as it receives the highest probability. The \textit{base pose} policy is then used in Layer 2 to determine the base pose for grasping object $o_1$.} 
    \label{fig:architecture}
\end{figure*}

\section{\name}
\label{sec:pa}
We decompose the task of learning to plan a base pose sequence into two sub-tasks: 
\begin{enumerate}
    \item Selecting the next object $\obj{m}$ to be grasped from among the $M$ remaining objects using the \textit{grasp sequence} policy $\pseq$ (Layer 1 in Fig.~\ref{fig:architecture}).
    
    \item Determining base pose $\abase{n}$ for grasping the selected object $\obj{m}$ using the \textit{base pose} policy $\pbase$ (Layer 2 in Fig.~\ref{fig:architecture}).
\end{enumerate}
Both sub-tasks are learned within the LL framework as shown in Fig.~\ref{fig:architecture}. The \textit{base pose} policy $\pbase$ is learned before the \textit{grasp sequence} policy $\pseq$ as $\abase{n}$ is required to perform the action predicted by $\pseq$. In the following sections, we describe learning to estimate the base pose for grasping (Section~\ref{ss:pa:bpp}) and grasping sequence (Section~\ref{ss:pa:gsp}).



\subsection{Learning to estimate base pose for grasping}
\label{ss:pa:bpp}
The \textit{base pose} policy $\pbase$ is learned using the Soft Actor-Critic (SAC) algorithm \cite{haarnoja2018soft} as a single-step policy in Layer 2 (see Fig. \ref{fig:architecture}(b)). Each episode consists of a single step wherein the object $\obj{m}$ is randomly placed on the table and the robot is randomly placed in the room within the 3m radius of the table. Given the object pose $\sobj{m}$ and the robot base pose $\srobot$ the agent learns to predict the base pose $\abase{n}$ for grasping the object $\obj{m}$:
\begin{equation}
    \abase{n} \sim \pbase(\ \cdot\ |\srobot, \sobj{m};\ \phi_{\text{base}}),
\end{equation}
where $\phi_{\text{base}}$ are learnable parameters. The base pose $\abase{n}$ is predicted in the object $\obj{m}$ frame; i.e., it is a transformation from the object frame $\obj{m}$ to the robot base frame $b$; $\transform{\obj{m}}{b}$. 

The  reward is defined as:
\begin{equation}
\begin{aligned}[t]
\rbase(\sobj{m},\abase{n}) = \phantom{+} \gamma_1  \cdot \mathbbm{1} (\textrm{collision}(\sobj{m},\abase{n})) +  \hspace*{1.5cm} \\
    \hspace*{0.5cm} \mathbbm{1}(\text{IK}(\sobj{m}, \abase{n})) \cdot  
    \left[ \gamma_2 + \frac{\gamma_3}{1+\tnav} + \frac{\gamma_4}{1+\tgrasp} \right],
\end{aligned}
\end{equation}
where $\mathbbm{1}(\textrm{collision}(\sobj{m},\abase{n}))$ is 1 if there is a collision with the table after moving to $\abase{n}$ and 0 otherwise; $\mathbbm{1}(\text{IK}(\sobj{i}, \abase{n}))$ is 1 if IK solutions are available to grasp the object $\obj{m}$ after moving to the base pose $\abase{n}$ and 0 otherwise; $\tnav$ is the time required to navigate from current robot base pose to the next base pose $\abase{n}$; $\tgrasp$ is the time required to grasp the object $m$ from the base pose $\abase{n}$ and $\gamma_1$, $\gamma_2$, $\gamma_3$, and $\gamma_4$ are hyper-parameters. 



\subsection{Learning to estimate grasping sequence}
\label{ss:pa:gsp}
In Layer 1, a probability for grasping each object $\obj{m}$ (among the $M$ remaining objects on the table) is learned while using the policy $\pbase$ already learned in Layer 2 for taking action $\abase{n}$. Thus, the agent uses a composite policy for exploration, and only the parameters $\phi_{\text{seq}}$ of $\pseq$ are learned here (see Fig. \ref{fig:architecture} (a)):
\begin{align}
\aobj{i} \sim & \pseq(\ \cdot\ | \sobj{i},  \{ \sobj{j} \}_j, \srobot; \phi_{\text{seq}}), \quad i,j \in \{1...M\} \land j \neq i , \nonumber \\
k = & \argmax_{i} \{\aobj{i}\}_i,  \\
\abase{n} \sim & \pbase(\ \cdot\ |\srobot, \sobj{k}; \phi_{\text{base}}), \nonumber 
\end{align}
where $\obj{i}$ is the object for which the grasp probability is being calculated, $\{ \sobj{j} \}_j$ are other objects in the scene that need to be grasped and $\phi_{\text{seq}}$ are learnable parameters. Use of the already learned \textit{base pose} policy reduces the exploration space as only the grasp sequence $\aseq$ order needs to be explored such that the reward over the entire episode is maximized. The reward for learning grasping sequence, $\rseq$, is calculated as:
\begin{equation}
\rseq(\StateS, \ActionS) = -\gamma_5 \cdot \tnav,
\end{equation}
where $\tnav$ is the time required to navigate from the current robot base pose to the predicted base pose and $\gamma_5$ is a hyper-parameter.

As the number of objects in the scene is not fixed, the state for learning the \textit{grasp sequence} policy $\pseq$ cannot be represented using a fixed-dimensional vector similar to the state for the \textit{base pose} policy $\pbase$. 
Since Graph Neural Networks (GNN) are invariant to node permutations, we use GNN for encoding state into fixed dimensional vector. 

\begin{figure}[h]
    \centering
    \includegraphics[width=0.98\columnwidth]{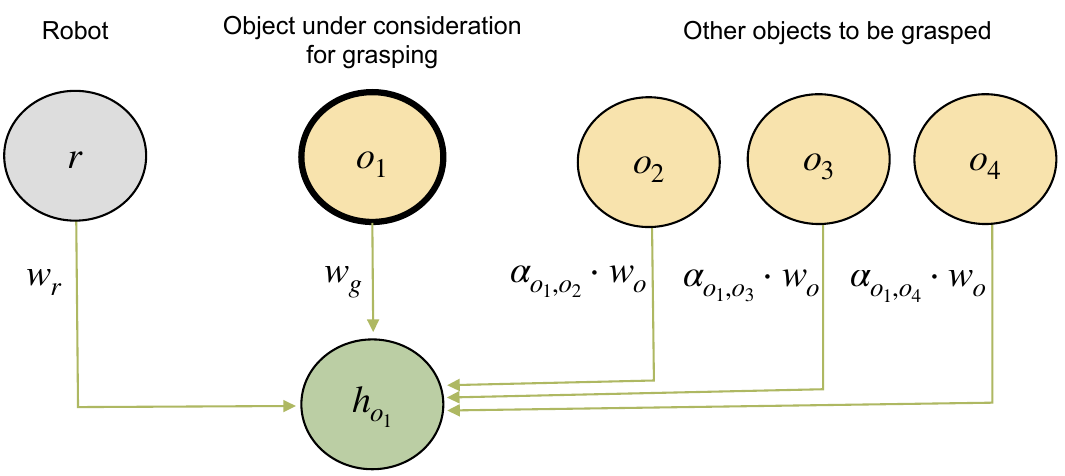}
    \caption{Attention-Based Graph Encoder.}
    \label{fig:encoder}
\end{figure}

\textbf{Encoder.} We use Graph Attention Layers \cite{velivckovic2018graph} to encode relevant information into a context embedding and formulate the grasp sequence policy $\pseq$ as a graph node regression problem. We use a heterogeneous graph with three different types of nodes: the robot $\robot$, the object under consideration for grasping $\obj{i}$, and other objects to be grasped $\obj{j}$. In the first layer, context embeddings for each node are generated as shown in Fig.~\ref{fig:encoder}. All three types of nodes $\robot$, $\obj{i}$, $\obj{j}$ are initially projected to a higher dimensional space using weights $w_r$, $w_g$, and $w_o$ respectively. Attention coefficients $\alpha$ are then calculated to determine the level of attention to be given to other objects in the scene $\obj{j}$ while encoding the context embedding for the object under consideration for grasping $\sobj{i}$. 
Thus, the context embedding for the object under consideration for grasping $\obj{i}$ is encoded as:
\begin{equation}
    h_{\obj{i}} = w_g \cdot \sobj{i} + w_r \cdot \srobot + \sum_{j \in \Omega(\obj{i})}  \alpha_{\obj{i}, \obj{j}} \cdot w_o \cdot \sobj{j},
\end{equation}
where $\Omega(\obj{i})$ are the object neighbors of the object $\obj{i}$ (other objects to be grasped in Fig.~\ref{fig:encoder}).

\begin{figure*}[h]
    \centering
    \includegraphics[width=17.5cm]{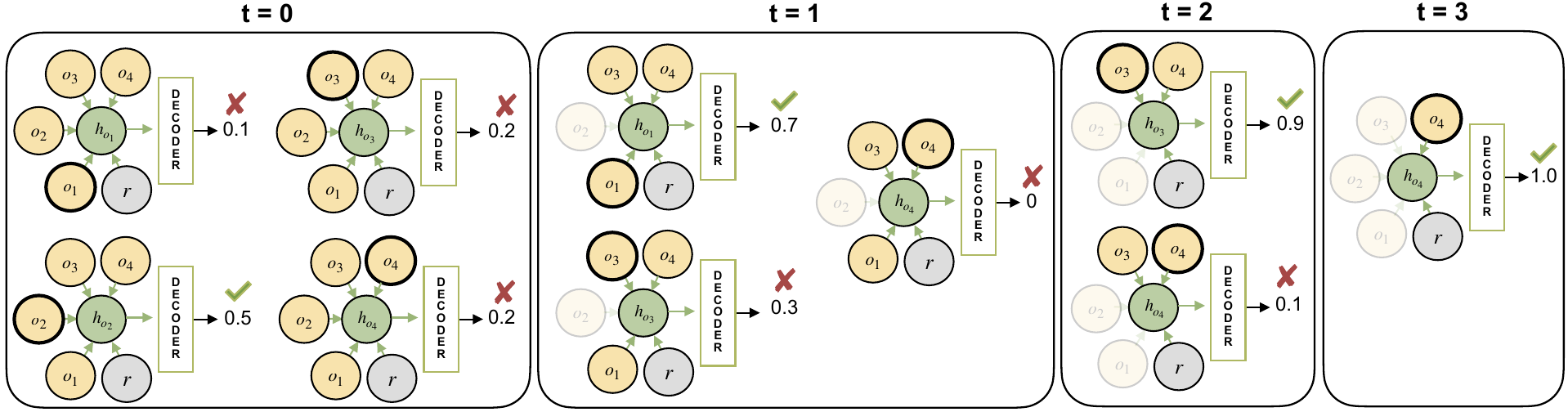}
    \caption{\textbf{Decoder:} The decoder takes the object context embeddings provided by the encoder as input and outputs the grasp probability for each object. Already grasped objects are masked during subsequent time steps. The example demonstrates the construction of a grasp sequence for a scene with four objects.}
    \label{fig:decoder}
\end{figure*}

\textbf{Decoder.} Each episode consists of $N$ steps (number of objects to grasp). At each time step, the grasp probability $\aobj{i}$ is calculated for all objects in the scene that have not yet been grasped. First, the encoder processes relevant information into a fixed-dimensional context embedding for each object. Subsequently, the decoder, which is a Multi-Layer Perceptron (MLP), predicts the grasp probability for each object using the encoded embedding (Graph Node Regression). The object to be grasped is selected by sampling from a categorical distribution parameterized by the grasp probabilities $\aobj{i}$ over objects. 
The base pose policy for the selected object class is used to determine the base pose and complete the pickup. Fig.~\ref{fig:decoder} presents a decoding example for a 4 objects scene.
The grasp sequence policy $\pseq$ is learned using REINFORCE \cite{sutton2018reinforcement} with a Greedy Rollout Baseline similar to \cite{kool2018attention}. The gradient of loss $\mathcal{L}(\phi_{\text{seq}}|S)$ for optimizing learnable parameters $\phi_{\text{seq}}$ in state $S$ is calculated as:
\begin{equation}
    \nabla \mathcal{L}(\phi_{\text{seq}}|\StateS) = - (\rseq(\StateS, \ActionS) - \rbaseline(\StateS)) \cdot \nabla\log \aobj{k},
\end{equation}
where $\aobj{k}$ is the grasp probability for the selected object $\obj{k}$ and $\rbaseline(S)$ is the baseline reward \cite{kool2018attention}. The baseline reduces variance and accelerates learning. Attempting to learn $\pseq$ using actor-critic proved unsuccessful due to difficulties in effectively representing the state and actions for learning the value function (critic).

\section{EXPERIMENTAL SETUP}
\label{sec:ee}

\subsection{Experiment and implementation Details}
\label{ss:es:ei}
For evaluating our work, we created an environment in NVIDIA Isaac Sim, using the mobile manipulator platform and a rectangular table (2.0m$\times$0.8m) with up to 10 YCB benchmark \cite{calli2015benchmarking} objects (belonging to 5 different classes) on it, as shown in Fig.~\ref{fig:idea}. Overhead (top-down) grasp poses for grasping objects of different classes were pre-defined.

For the \textit{base pose} policy $\pbase$ learning, we used a Multi-Layer Perceptron (MLP) with three hidden layers in both the Actor and Critic networks in SAC \cite{haarnoja2018soft}. Each hidden layer comprised 256 neurons with ReLU activation. The learning rate was set to 3e-4. The reward hyper-parameters $\gamma_1$, $\gamma_2$, $\gamma_3$ and $\gamma_4$, were empirically set to -2e5, 1e6, 5e5, and 5e5, respectively.

For the \textit{grasp sequence} policy $\pseq$ learning, we used a 64-bit vector representation learned through five 64-bit Graph Attention Layers \cite{velivckovic2018graph} followed by ReLU activation. This vector representation served as input to an MLP with two hidden layers, each comprising 64 neurons with ReLU activation. Both networks were optimized simultaneously using an Adam optimizer \cite{kingma2014adam} with gradients computed via REINFORCE with a greedy rollout baseline \cite{kool2018attention} and a learning rate of 1e-3. The reward hyper-parameter $\gamma_5$ was empirically set to 1e3.

In each episode, the robot's starting pose was randomly sampled within a 2.5-3m radius around the table. For an episode with $N$ objects on the table, the robot can select up to $N$ base poses for grasping all the objects. The episode terminates when the robot has visited $N$ base poses or has grasped all the objects or the predicted base pose cannot be reached as it will result in a collision with the table. The state $\StateS$ was calculated based on the states provided by the simulator. 

To expedite training, instead of using a navigation stack to move to the predicted base poses, the robot was teleported. The navigation cost was computed using the approximation based on the linear and angular travel distance \cite{wang2021optimal}. 
For grasping, the grasp execution time was determined using the Lula Trajectory generator available in NVIDIA Isaac Sim, which provides an approximate grasp trajectory execution time without executing the actual trajectory, while IRMs were computed using the Lula Kinematics Solver with a discretization of 10cm and 45$^\circ$. For the algorithmic implementation of our approach, we used the Mushroom RL library \cite{carlo2021}.


All the experiments were carried out on the workstation equipped with Intel Core i9-13900KF 24-Core processor, 64 GB RAM, and NVIDIA GeForce RTX 4090 24GB GPU.

\begin{table*}[!hbt]
\resizebox{\textwidth}{!}{%
\begin{tabular}{@{}cccclclclclcc@{}}
\toprule
\multirow{2}{*}{Method} &
  \multicolumn{2}{c}{\multirow{2}{*}{\begin{tabular}[c]{@{}c@{}}\% of objects \\ grasped\end{tabular}}} &
  \multicolumn{2}{c}{\multirow{2}{*}{\begin{tabular}[c]{@{}c@{}}Planning time (in s)\\ \\ P\end{tabular}}} &
  \multicolumn{2}{c}{\multirow{2}{*}{\begin{tabular}[c]{@{}c@{}}Navigation time (in s)\\ \\ N\end{tabular}}} &
  \multicolumn{2}{c}{\multirow{2}{*}{\begin{tabular}[c]{@{}c@{}}Grasping time (in s)\\ \\ G\end{tabular}}} &
  \multicolumn{2}{c}{\multirow{2}{*}{\begin{tabular}[c]{@{}c@{}}Total execution\\ time (in s)\\ N + G\end{tabular}}} &
  \multicolumn{2}{c}{\multirow{2}{*}{\begin{tabular}[c]{@{}c@{}}Total planning and\\ execution time (in s)\\ P+N+G\end{tabular}}} \\
 &
  \multicolumn{2}{c}{} &
  \multicolumn{2}{c}{} &
  \multicolumn{2}{c}{} &
  \multicolumn{2}{c}{} &
  \multicolumn{2}{c}{} &
  \multicolumn{2}{c}{} \\
\multicolumn{1}{l}{} &
  \multicolumn{1}{l}{} &
  \multicolumn{1}{l}{} &
  \multicolumn{1}{l}{} &
   &
  \multicolumn{1}{l}{} &
   &
  \multicolumn{1}{l}{} &
   &
  \multicolumn{1}{l}{} &
   &
  \multicolumn{1}{l}{} &
  \multicolumn{1}{l}{} \\ \cmidrule(l){2-13} 
 &
  5-objs &
  10-objs &
  5-objs &
  \multicolumn{1}{c}{10-objs} &
  5-objs &
  \multicolumn{1}{c}{10-objs} &
  5-objs &
  \multicolumn{1}{c}{10-objs} &
  5-objs &
  \multicolumn{1}{c}{10-objs} &
  5-objs &
  10-objs \\ \midrule
  \textit{Approximate Methods} & \multicolumn{12}{c}{}\\
PBG &
  \textcolor{red}{10.0} &
  \textcolor{red}{$\phantom{0}$8.9} &
  $\phantom{00}$\textbf{0.0}$\pm$\textbf{0.0} &
  $\phantom{00}$\textbf{0.0}$\pm$\textbf{0.0} &
  \textcolor{blue}{\textit{12.8$\pm$4.7}} &
  \textcolor{blue}{\textit{15.7$\pm$6.7}} &
  \textcolor{blue}{\textit{$\phantom{0}$7.8$\pm$12.2}} &
  \textcolor{blue}{\textit{$\phantom{0}$15.3$\pm$16.2}} &
  \textcolor{blue}{\textit{$\phantom{0}$20.7}} &
  \textcolor{blue}{\textit{$\phantom{0}$31.0}} &
  \textcolor{blue}{\textit{$\phantom{0}$20.7}} &
  \textcolor{blue}{\textit{$\phantom{0}$31.0}} \\
PBG-GC &
  94.4 &
  89.5 &
  136.0$\pm$2.5 &
  273.2$\pm$4.0 &
  21.3$\pm$6.2 &
  28.8$\pm$7.1 &
  72.9$\pm$12.0 &
  137.2$\pm$18.6 &
  $\phantom{0}$94.2 &
  166.0 &
  230.3 &
  439.3 \\
MBP &
  \textcolor{red}{39.6} &
  \textcolor{red}{53.4} &
  $\phantom{00}$\textbf{0.0}$\pm$\textbf{0.0} &
  $\phantom{00}$\textbf{0.0}$\pm$\textbf{0.0} &
  \textcolor{blue}{\textit{12.0$\pm$5.9}} &
  \textcolor{blue}{\textit{17.5$\pm$5.2}} &
  \textcolor{blue}{\textit{29.9$\pm$14.9}} &
  \textcolor{blue}{\textit{$\phantom{0}$80.7$\pm$22.7}} &
  \textcolor{blue}{\textit{$\phantom{0}$41.9}} &
  \textcolor{blue}{\textit{$\phantom{0}$98.2}} &
  \textcolor{blue}{\textit{$\phantom{0}$41.9}} &
  \textcolor{blue}{\textit{$\phantom{0}$98.2}} \\
MBP-GC &
  96.0 &
  \textbf{97.9} &
  129.1$\pm$1.9 &
  250.5$\pm$2.7 &
  \textbf{19.3$\pm$5.4} &
  \textbf{18.3$\pm$2.8} &
  74.0$\pm$$\phantom{0}$6.8 &
  151.2$\pm$$\phantom{0}$9.3 &
  $\phantom{0}$93.3 &
  169.6 &
  222.5 &
  420.1 \\
  \midrule
  \textit{Exact Methods} & \multicolumn{12}{c}{}\\
DP &
  \textbf{98.4} &
  97.1 &
  131.4$\pm$1.7 &
  259.9$\pm$3.5 &
  22.3$\pm$7.0 &
  19.4$\pm$3.4 &
  \textbf{65.9$\pm$$\phantom{0}$4.8} &
  \textbf{127.1$\pm$$\phantom{0}$6.8} &
  \textbf{$\phantom{0}$88.3} &
  \textbf{146.4} &
  219.7 &
  406.3 \\ \midrule
\name~(\textbf{Ours}) &
  97.5 &
  94.8 &
  $\phantom{00}$\textbf{0.0}$\pm$\textbf{0.0} &
  $\phantom{00}$\textbf{0.0}$\pm$\textbf{0.0} &
  27.8$\pm$6.0 &
  36.2$\pm$4.6 &
  76.0$\pm$$\phantom{0}$8.0 &
  148.6$\pm$$\phantom{0}$9.9 &
  $\phantom{0}$103.8 &
  184.8 &
  \textbf{103.8} &
  \textbf{184.8} \\ \bottomrule
\end{tabular}%
}
\caption{Success rate \& time comparison. \textcolor{red}{Red} highlights unacceptable success rates \& hence \textcolor{blue}{\textit{blue italic values}} are irrelevant.}
\label{tab:quant_results}
\end{table*}

\subsection{Experiment objectives and baselines}
\label{ss::ee:eob}
The experiments aim to verify whether the proposed learning-based method can produce solutions comparable to those obtained using exact and approximate methods in a shorter computation time (Section~\ref{ss:res:time_perf}). The following baselines are considered:
\begin{itemize}[leftmargin=*]
    \setlength\itemsep{0.25em}
    \item[] \textbf{Proximity-Based Greedy selection (PBG)}: IRMs are used to obtain a set of base poses from which each object can be grasped. The robot selects the object closest to it for grasping next and uses the greedy selection strategy based on the navigation cost to select the base pose for grasping.
    \item[] \textbf{Minimum Base Poses (MBP)}: IRMs are used to obtain a set of base poses from which each object can be grasped. The base poses are then selected to minimize the number of base poses required for grasping all objects. Subsequently, the robot employs a greedy selection strategy based on navigation cost to determine the next base pose. This is similar to \cite{harada2015base, xu2020planning} without making any assumptions regarding the action space such as fixed base orientation.
    \item[] \textbf{Dynamic Programming (DP)}: IRMs are used to acquire a set of base poses from which each object can be grasped. Then navigation and grasp execution costs are computed for all the base poses. Dynamic Programming with memoization is then used to plan the optimal sequence of base poses similar to \cite{sorensen2024planning}. 
\end{itemize}
PBG and MBP are approximate methods, whereas DP is an exact method. As IRM does not guarantee that a grasp trajectory can be planned from the base pose, we also present results for PBG and MBP by first validating the set of selected base poses and only considering the base poses from which trajectory can be planned for grasping the object. These baselines are referred to as \textbf{PBG-GC} and \textbf{MBP-GC}. 

In addition, we present ablation studies in Section~\ref{ss:res:learn_perf} to validate our design choices. First, we compare the learning performance of \textit{base pose} policy $\pbase$ when the base pose $\abase{n}$ is predicted in the frame of the object $\obj{m}$ selected for grasping and in the table frame $\world$. Second, we compare the learning performance of \textit{grasp sequence} policy $\pseq$ (i) with and without using a greedy rollout baseline with REINFORCE and (ii) with and without using attention coefficients $\alpha$ for encoding context embedding.


\section{RESULTS}
\label{sec:res}

\subsection{Experiment 1: Planning and execution time analysis}
\label{ss:res:time_perf}
In Table~\ref{tab:quant_results}, we present the mean and standard deviation values for \textit{planning time}, \textit{execution time} (\textit{navigation} $\tnav$ and \textit{grasping} $\tgrasp$), \textit{total time}, and the \textit{percentage of objects grasped} for the five selected baselines and \name. We considered two tasks: `5-objs' and `10-objs', each with 5 and 10 objects to grasp, respectively. We evaluated each task over 50 random scenes. During navigation, the maximum base linear and angular velocities were set to 0.5m/s and 0.5rad/s. During manipulation with the UR5e manipulator, the maximum velocities for shoulder and elbow joints were set to 1.0rad/s, and for wrist joints, it was set to 2.0rad/s.

In real-world setups, several challenges related to the robot's perception can contribute to additional execution time and failures. These include planning robot camera views to accurately estimate object poses before grasping, performing 6D pose estimation, etc. Furthermore, inaccurate pose estimates can lead to grasp failures. Since these challenges are not addressed in this work, to avoid their influence, we used a simulated environment for evaluation.

Table~\ref{tab:quant_results} shows that, as expected, DP, being an exact method, produces the most optimal solutions in terms of total \textit{execution time}, with more than 97\% of objects successfully grasped. PBG-GC and MBP-GC, which are approximate methods, also produce near-optimal solutions. However, all these baselines have very high \textit{planning} time. The majority of the \textit{planning} time is attributed to the computation of navigation and grasping costs for the action space indicated by the IRM. PBP and MBP have very low \textit{planning} time because they do not involve any cost computation, assuming that trajectories can be planned to grasp the object from all the base poses in the IRM. However, as this assumption doesn't always hold true, especially for 6 DOF manipulators like UR5e, they tend to perform poorly.

\name$ $ produces solutions comparable to those produced by PBG-GC and MBP-GC in terms of total \textit{execution time} and the \textit{percentage of objects grasped}, but with significantly lower \textit{planning time}. While DP produces better solutions in terms of \textit{execution time}, it also has high \textit{planning time}. Therefore, when it comes to total \textit{planning and execution time} \name~ outperforms all the baselines for both the tasks.

\subsection{Experiment 2: Qualitative results} 
\begin{figure*}[h]
    \centering
    \includegraphics[width=18.0cm]{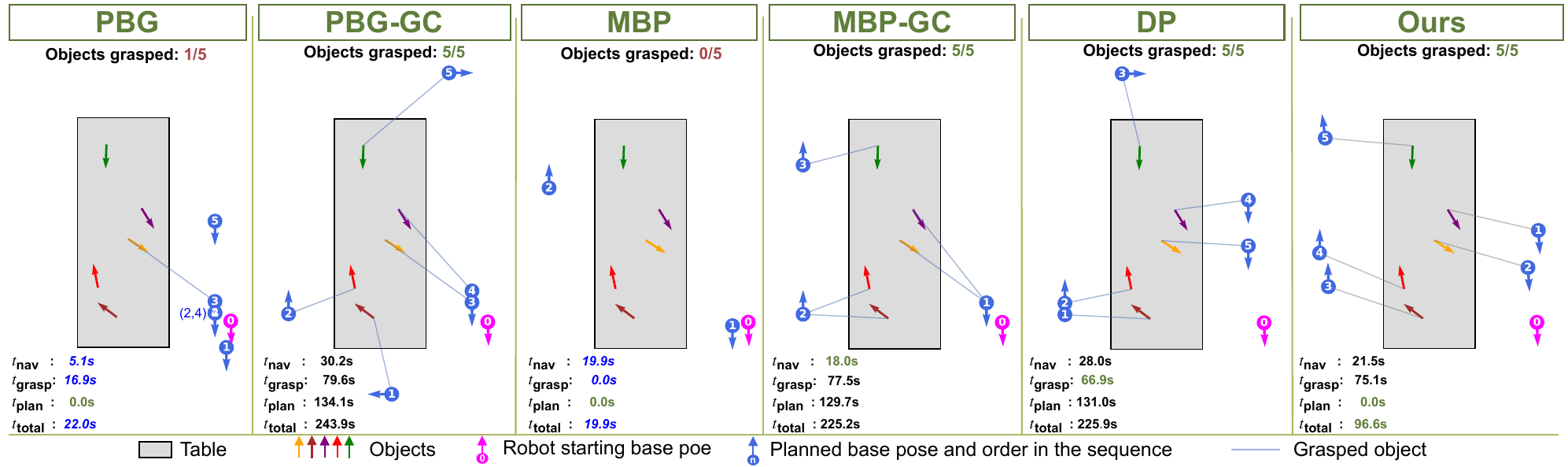}
    \caption{Sequence of base poses planned by baselines and our method for a random scene. \textcolor{red}{Red} highlights unacceptable success rates (objects grasped) \& hence \textcolor{blue}{\textit{blue italic values}} are irrelevant. Predicted base poses are centers of the mobile base. The manipulator is positioned on the right back corner of the mobile base as can be seen in Fig.~\ref{fig:idea}.} \label{fig:qual_results}
\end{figure*}

In Fig.~\ref{fig:qual_results}, we present qualitative results for a random scene using all the baselines and our method \name. It can be observed that the base poses $\abase{n}$ planned by \name~are farther away from the table compared to the DP solutions. This occurs because the \textit{base pose} policy $\pbase$ learns to maintain a safe distance from the table, given the high penalty for collision with the table. Consequently, \name~has longer \textit{grasping time} as the manipulator requires longer trajectories to grasp the objects.

All baselines have a discrete base pose space with the discretization of 10cm and 45$^\circ$. Consequently, the base poses planned by the baselines are perfectly aligned with each other, requiring only linear robot motions to move between them. In contrast, \name~predicts base poses in the continuous space and hence they are not perfectly aligned. As a result, the robot requires both linear and angular motion to move between them. This incurs significant navigation costs and explains the higher navigation times for \name.

In this work, we have used generic rewards. Both of the above issues can be addressed through task and robot-specific reward engineering. 




\subsection{Experiment 3: Ablation analysis}
\label{ss:res:learn_perf}

\begin{figure}[ht]
\begin{minipage}[b]{0.48\linewidth}
\centering
\includegraphics[width=\textwidth]{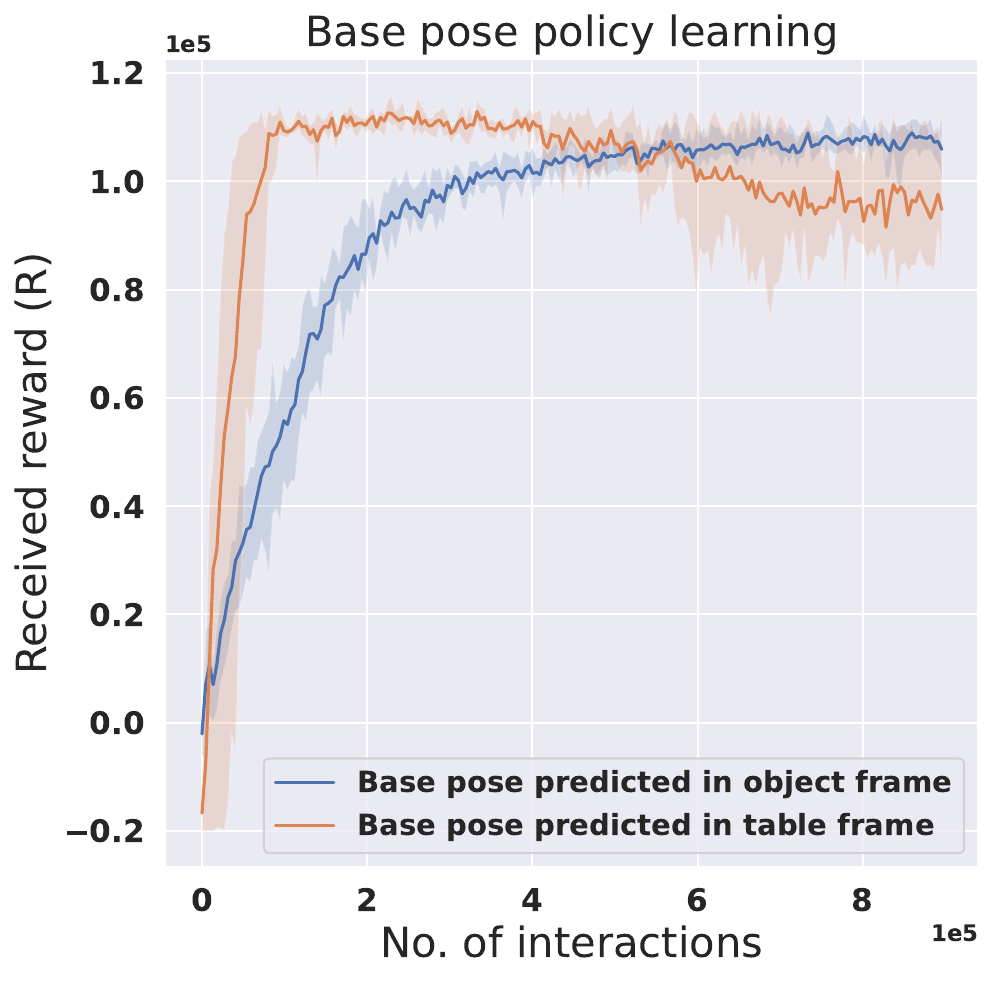}
\subcaption{\textit{Base pose} policy $\pbase$}
\label{fig:bp_learning}
\end{minipage}
\hspace{0.0cm}
\begin{minipage}[b]{0.48\linewidth}
\centering
\includegraphics[width=\textwidth]{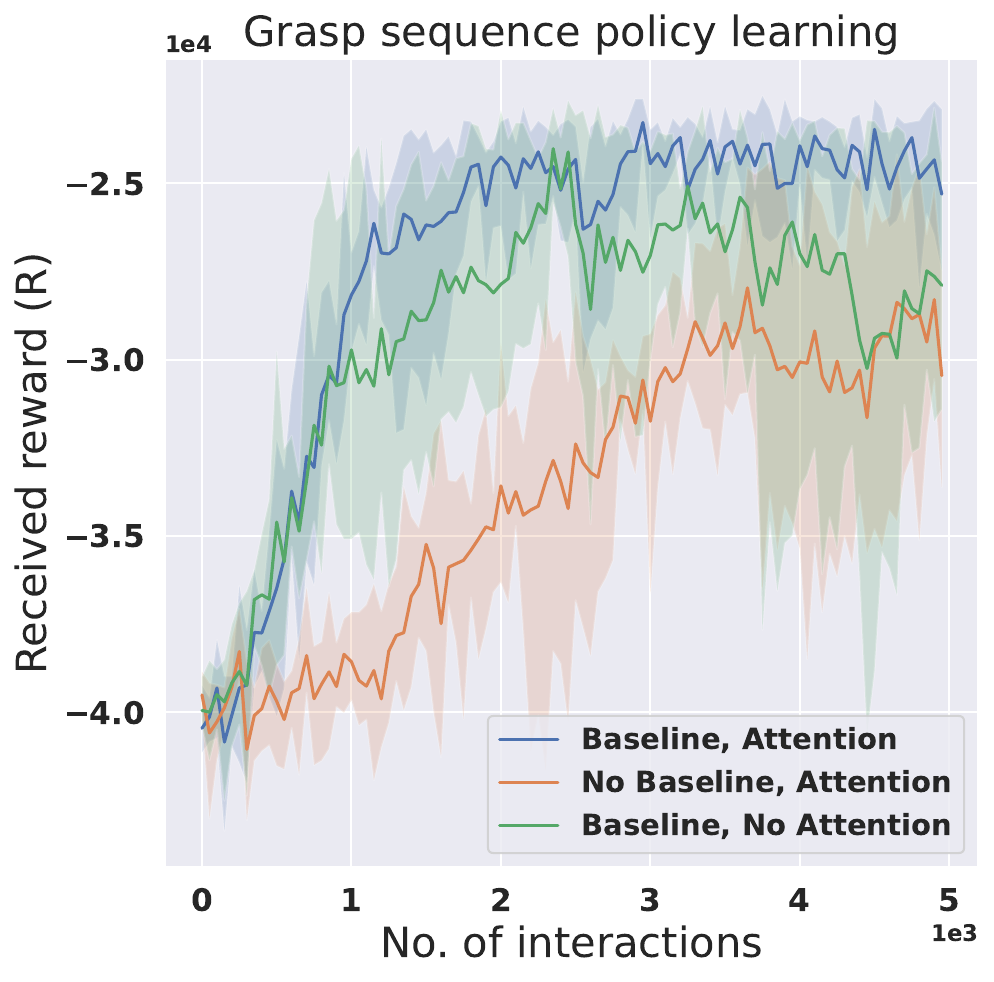}
\subcaption{\textit{Grasp sequence} policy $\pseq$}
\label{fig:gs_learning}
\end{minipage}
\caption{Comparing the learning performance of \textit{base pose} policy $\pbase$ and \textit{grasp sequence} policy $\pseq$ under different scenarios. Figures show mean over 5 seed runs and min and max variations.}
\label{fig:learning}
\end{figure}

\begin{figure}[ht]
\includegraphics[width=\linewidth]{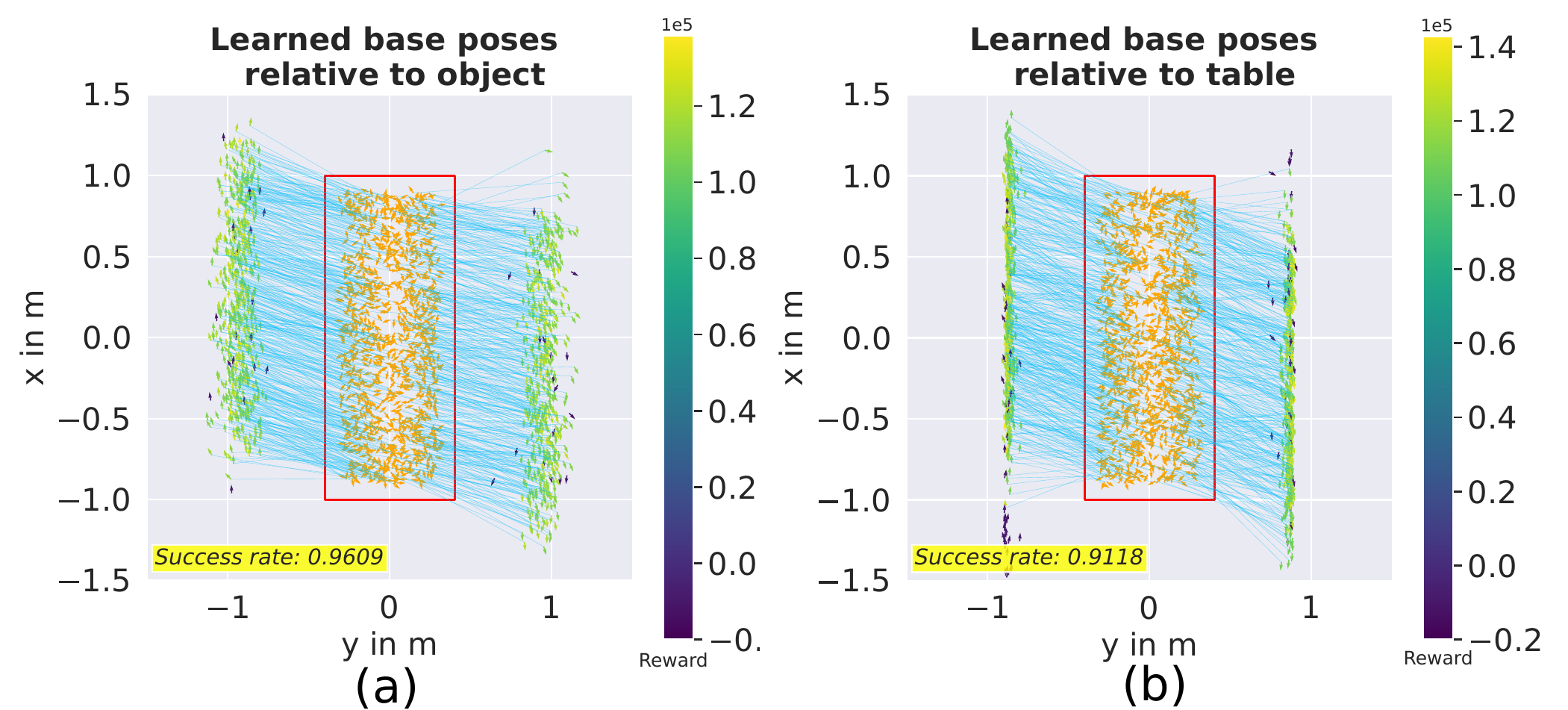}
\caption{Learned base poses (colored with colorbar) in \textbf{(a)} object frame and \textbf{(b)} table frame. The red rectangle denotes the table, objects are orange, and blue lines connect each base pose to the grasped object.}
\label{fig:bp_space}
\end{figure}

\noindent\textbf{Learning performance of $\pbase$.}
Fig.~\ref{fig:bp_learning} compares the performance of \textit{base pose} policy $\pbase$ using the object frame $\obj{m}$ and the table frame $\world$ for predicting base poses $\abase{n}$ during training. When using the table frame for predicting base poses, $\pbase$ quickly learns highly optimal base poses for grasping objects in specific regions of the table. However, it fails to effectively explore the base pose space for objects placed anywhere on the table. Conversely, the use of object frame leads to very stable learning. This observation is further supported by Fig.~\ref{fig:bp_space}, which shows the learned base poses (colorbar colors) in both cases for 1000 random object poses (orange) on the table (red rectangle). It can be seen that base poses learned in object frame are more generic and have a grasp success rate of over 96\%, compared to base poses learned in the table frame, which achieve only around a 91\% grasp success rate. These results can be attributed to the fact that in the object frame, the agent only needs to explore the region around the object to predict base poses, simplifying the learning process.

\noindent\textbf{Learning performance of $\pseq$.}
Fig.~\ref{fig:gs_learning} compares the performance of the \textit{grasp sequence} policy $\pseq$ with and without the Greedy Rollout Baseline and the use of attention coefficients $\alpha$ during training. The use of the baseline accelerates learning and results in better policies with higher rewards. This improvement occurs because the baseline reduces variance during learning \cite{kool2018attention}. Additionally, we observe that learning attention coefficients (blue) leads to stable learning compared to learning without attention coefficients (green). The attention coefficients learn to determine how much attention should be given to other objects in the scene, thus generating more informative context embeddings.

\section{CONCLUSION AND FUTURE WORK}
\label{sec:con_and_fut}
In this work, we have presented \name, a learning-based approach for planning mobile manipulator base pose sequences for pick-up tasks while optimizing total navigation and grasping time. We compared our work with three baselines (+2 variations) that use exact and approximate methods for solving the problem.  Our experiments show that \name$ $ produces comparable solutions in significantly less computation time. 
In this way, \name~ allows the robots to quickly re-plan when the object configuration in the scene changes, either due to human intervention or collision with other objects in the scene. 

A limitation of this work is it doesn't consider uncertainty in the object poses and the robot's self-localization. However, execution failures can be prevented by estimating the uncertainties \cite{naik2022multi} and assessing whether the estimated errors are acceptable for the successful execution of the planned action \cite{naik2024c}. If uncertainties exceed acceptable thresholds, the robot can defer the action execution until uncertainties reduce to acceptable levels. Future works should investigate the inclusion of pose uncertainties in the state space during learning so that robots can plan the base poses considering uncertainties.

\addtolength{\textheight}{-0cm}   





\section*{ACKNOWLEDGMENT}
This work was supported by the Innovation Fund Denmark's FacilityCobot project and the European Union's Fluently project.


\bibliographystyle{IEEEtran}
\typeout{}
\bibliography{references}

\end{document}